\documentclass[sigconf]{aamas}

\usepackage{balance} % for balancing columns on the final page

\setcopyright{ifaamas}
\acmConference[AAMAS '23]{Proc.\@ of the 22nd International Conference
on Autonomous Agents and Multiagent Systems (AAMAS 2023)}{May 29 -- June 2, 2023}
{London, United Kingdom}{A.~Ricci, W.~Yeoh, N.~Agmon, B.~An (eds.)}
\copyrightyear{2023}
\acmYear{2023}
\acmDOI{}
\acmPrice{}
\acmISBN{}

\title{Learning Reward Machines in Cooperative Multi-Agent Tasks}

\author{Leo Ardon}
\affiliation{%
  \institution{Imperial College London}
  \country{}
}

\author{Daniel Furelos-Blanco}
\affiliation{%
  \institution{Imperial College London}
  \country{}
}
\author{Alessandra Russo}
\affiliation{%
  \institution{Imperial College London}
  \country{}
}

\begin{abstract}

This paper presents a novel approach to Multi-Agent Reinforcement Learning (MARL) that combines cooperative task decomposition with the learning of reward machines (RMs) encoding the structure of the sub-tasks. 
The proposed method helps deal with the non-Markovian nature of the rewards in partially observable environments and improves the interpretability of the learnt policies required to complete the cooperative task.
The RMs associated with each sub-task are learnt in a decentralised manner and then used to guide the behaviour of each agent.
By doing so, the complexity of a cooperative multi-agent problem is reduced, allowing for more effective learning.
The results suggest that our approach is a promising direction for future research in MARL, especially in complex environments with large state spaces and multiple agents.

\end{abstract}


\newcommand{\BibTeX}{\rm B\kern-.05em{\sc i\kern-.025em b}\kern-.08em\TeX}

\usepackage{caption}
\usepackage{subcaption}
\usepackage{float}
\usepackage{xcolor}
\graphicspath{ {./figures/} }

\usepackage{amsmath}
\usepackage{multirow}
\usepackage{pifont}
\usepackage{tikz}
\usepackage{xspace}
\usepackage[linesnumbered,noend]{algorithm2e}
\usepackage{bm}

\usetikzlibrary{automata, positioning, shapes, arrows, shapes.geometric}

\newcommand{\rmname}{M}
\newcommand{\rmstates}{\mathcal{U}}
\newcommand{\propositions}{\mathcal{P}}
\newcommand{\rmdeltau}{\delta_u}
\newcommand{\rmdeltar}{\delta_r}
\newcommand{\rmstate}{u}
\newcommand{\rminitstate}{\rmstate_0}
\newcommand{\rmstatefinal}{\rmstate_A}

\newcommand{\mdpstates}{\mathcal{S}}
\newcommand{\mdpactions}{\mathcal{A}}
\newcommand{\mdptransition}{p}
\newcommand{\mdpreward}{r}
\newcommand{\mdpdisc}{\gamma}
\newcommand{\mdplfunc}{l}
\newcommand{\mdpterm}{\tau}
\newcommand{\policy}{\pi}

\newcommand{\sahistory}{h}
\newcommand{\ltrace}{\lambda}
\newcommand{\lgtrace}{\lambda^{\top}}
\newcommand{\litrace}{\lambda^{\bot}}
\newcommand{\proplabel}{\mathcal{L}}
\newcommand{\traceset}{\Lambda}
\newcommand{\gtraceset}{\traceset_\top}
\newcommand{\itraceset}{\traceset_\bot}

\newcommand{\buttonsAgentTwo}{A_2}
\newcommand{\buttonsAgentThree}{A_3}
\newcommand{\redButton}{R_B}
\newcommand{\yellowButton}{Y_B}
\newcommand{\greenButton}{G_B}

\newcommand{\agentTwoRedButton}{\buttonsAgentTwo^{\redButton}}
\newcommand{\agentTwoNotRedButton}{\buttonsAgentTwo^{\lnot \redButton}}
\newcommand{\agentThreeRedButton}{\buttonsAgentThree^{\redButton}}
\newcommand{\agentThreeNotRedButton}{\buttonsAgentThree^{\lnot \redButton}}

\newcommand{\markovgame}{\bm{\mathcal{G}}}
\newcommand{\mgamestate}{\bm\mdpstates}
\newcommand{\mgameinitstate}{\bm{s}_I}
\newcommand{\mgameactions}{\bm\mdpactions}
\newcommand{\numagents}{N}

\newcommand{\qfunc}{q}

\newcommand{\threebuttons}{\textsc{ThreeButtons}\xspace}
\newcommand{\rendezvous}{\textsc{Rendezvous}\xspace}

\newcommand{\goal}{Goal}
\newcommand{\hDist}{1.5cm}
\newcommand{\vDist}{1.0cm}

\DeclareMathOperator{\codeif}{\mathtt{:-} }

\tikzset{auto,
    ->,
    >=stealth,
    node distance=2.2cm,
    every node/.style={scale=0.85, minimum size=0pt, inner sep=0pt}
}

\begin{document}

%%% The following commands remove the headers in your paper. For final 
%%% papers, these will be inserted during the pagination process.

\pagestyle{fancy}
\fancyhead{}

%%% The next command prints the information defined in the preamble.

\maketitle 

%%%%%%%%%%%%%%%%%%%%%%%%%%%%%%%%%%%%%%%%%%%%%%%%%%%%%%%%%%%%%%%%%%%%%%%%

\section{Introduction}

 With impressive advances in the past decade, the Reinforcement Learning (RL) \cite{SuttonB18} paradigm appears as a promising avenue in the quest to have machines learn autonomously how to achieve a goal. 
 Originally evaluated in the context of a single agent interacting with the rest of the world, researchers have since then extended the approach to the multi-agent setting (MARL), where multiple autonomous entities learn in the same environment.
 Multiple agents learning concurrently brings a set of challenges, such as partial observability, non-stationarity, and scalability; however, this setting is more realistic. Like in many real-life scenarios, the actions of one individual often affect the way others act and should therefore be considered in the learning process.

A sub-field of MARL called Cooperative Multi-Agent Reinforcement Learning (CMARL) focuses on learning policies for multiple agents that must coordinate their actions to achieve a shared objective. 
With a ``divide and conquer'' strategy, complex problems can thus be decomposed into smaller and simpler ones assigned to different agents acting in parallel in the environment.
The problem of learning an optimal sequence of actions now also includes the need to strategically divide the task among agents that must learn to coordinate and communicate in order to find the optimal way to accomplish the goal.

An emergent field of research in the RL community exploits finite-state machines to encode the structure of the reward function; ergo, the structure of the task at hand. 
This new construct, introduced by \citet{Icarte_Klassen_Valenzano_McIlraith_2018} and called \emph{reward machine} (RM), can model non-Markovian reward and provides a symbolic interpretation of the stages required to complete a task. 
While the structure of the task is sometimes known in advance, it is rarely true in practice. Learning the RM has thus been the object of several works in recent years \cite{ChristoffersenLTM20,FurelosBlancoLJBR22,Furelos_Blanco_Law_Jonsson_Broda_Russo_2021,HasanbeigJAMK21,IcarteWKVCM19,XuGAMNT020} as a way to reduce the human input.
In the multi-agent settings, however, the challenges highlighted before make the learning of the RM encoding the global task particularly difficult. Instead, it is more efficient to learn the RMs of smaller tasks obtained from an appropriate task decomposition. The ``global'' RM can later be reconstructed with a parallel composition of the RM of the sub-tasks.

Inspired by the work of \citet{Neary_Xu_Wu_Topcu_2021} using RMs to decompose and distribute a global task to a team of collaborative agents, we propose a method combining task decomposition with autonomous learning of the RMs. Our approach offers a more scalable and interpretable way to learn the structure of a complex task involving collaboration among multiple agents.
%
%Decomposing the global task into independently solvable sub-tasks addresses the non-stationarity associated with learning multiple policies simultaneously. 
%
%It also helps deal with the scalability issues faced while trying to lean the task structure of the global task by focusing on smaller and simpler sub-problems.
We present an algorithm that learns in parallel the RMs associated with the sub-task of each agent and their associated RL policies whilst guaranteeing that their executions achieve the global cooperative task. We experimentally show that with our method a team of agents is able to learn how to solve a collaborative task in two different environments.

\section{Background}
This section gives a summary of the relevant
background to make the paper self-contained. Given a finite set $\mathcal{X}$, $\Delta(\mathcal{X})$ denotes the probability simplex over $\mathcal{X}$, $\mathcal{X}^\ast$ denotes (possibly empty) sequences of elements from $\mathcal{X}$, and $\mathcal{X}^+$ is a non-empty sequence. The symbols $\bot$ and $\top$ denote false and true, respectively.

\subsection{Reinforcement Learning}
\label{sec:back_rl}
We consider \emph{$T$-episodic} \emph{labeled} Markov decision processes (MDPs)~\cite{XuGAMNT020}, characterized by a tuple $\langle \mdpstates, s_{I}, \mdpactions, \mdptransition, \mdpreward, T, \mdpdisc, \propositions, \mdplfunc \rangle$ consisting of a set of states $\mdpstates$, an initial state $s_{I}\in\mdpstates$, a set of actions $\mdpactions$, a transition function $\mdptransition:\mdpstates \times \mdpactions \to \Delta(\mdpstates)$, a \emph{not} necessarily Markovian reward function $\mdpreward:(\mdpstates \times \mdpactions)^+ \times \mdpstates \to \mathbb{R}$, the time horizon $T$ of each episode, 
%% a termination function $\mdpterm:(\mdpstates\times \mdpactions)^\ast\times\mdpstates\to \{\bot, \top\}$, 
%
a discount factor $\mdpdisc \in [0,1)$, a finite set of propositions $\propositions$, and a labeling function $\mdplfunc:\mdpstates \times \mdpstates \to 2^\propositions$. 

A (state-action) \emph{history} $\sahistory_t=\langle s_0, a_0, \ldots, s_t\rangle\in (\mdpstates \times \mdpactions)^\ast\times \mdpstates$ is mapped into a \emph{label trace} (or trace) $\ltrace_t=\langle\mdplfunc(s_0, s_1), \ldots,\mdplfunc(s_{t-1}, s_t) \rangle\in (2^\propositions)^+$ by applying the labeling function to each state transition in $\sahistory_t$. 
We assume that the reward
%and termination 
function can be written in terms of a label trace, i.e. formally $\mdpreward(\sahistory_{t+1})=\mdpreward(\ltrace_{t+1}, s_{t+1})$.
%and $\mdpterm(\sahistory_t)=\mdpterm(\ltrace_t,s_t)$
The goal is to find an optimal \emph{policy} $\policy:(2^\propositions)^+\times\mdpstates\to\mdpactions$, mapping (traces-states) to actions, in order to maximize the expected cumulative discounted reward $R_t=\mathbb{E}_\policy\left[\sum^T_{k=t}\mdpdisc^{k-t}\mdpreward(\ltrace_{k+1},s_{k+1})\right]$.

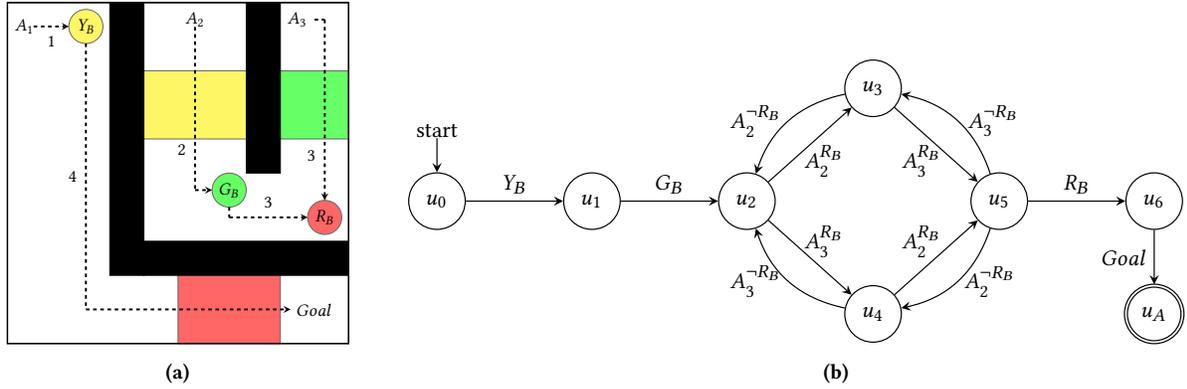
\begin{figure*}
    \captionsetup{justification=centering}
    \centering
    \begin{subfigure}{0.325\textwidth}
        \centering
        \resizebox{0.8\columnwidth}{!}{
            \begin{tikzpicture}[scale=1.0]
                \filldraw[fill=black!0!white] (0,0) rectangle (10,10);
                \filldraw[fill=black, draw=black] (3,2) rectangle (4,10);
                \filldraw[fill=black, draw=black] (4,2) rectangle (10,3);
                \filldraw[fill=black, draw=black] (7,5) rectangle (8,10);
                \filldraw[fill=yellow, opacity=0.6, draw=black] (4,6) rectangle (7,8);
                \filldraw[fill=green, opacity=0.6, draw=black] (8,6) rectangle (10,8);
                \filldraw[fill=red, opacity=0.6, draw=black] (5,0) rectangle (8,2);
            
                \filldraw[fill=yellow, opacity=0.6, draw=black] (2.3, 9.3) circle (0.5cm);
                \filldraw[fill=green, opacity=0.6, draw=black] (6.5, 4.5) circle (0.5cm);
                \filldraw[fill=red, opacity=0.6, draw=black] (9.3, 3.7) circle (0.5cm);
                \node at (2.3, 9.3) {\Huge $\yellowButton$};
                \node at (6.5, 4.5) {\Huge $\greenButton$};
                \node at (9.3, 3.7) {\Huge $\redButton$};
                
                \node[anchor=center] at (9.0, 1.0) (g) {\Huge $\goal$};
                \node[anchor=center] at (0.5, 9.3) (a1) {\Huge $A_1$};
                \node[anchor=center] at (5.5, 9.5) (a2) {\Huge $A_2$};
                \node[anchor=center] at (8.5, 9.5) (a3) {\Huge $A_3$};
                
                \draw[draw=black, very thick] (0,0) rectangle (10,10);
                \path[draw, dashed, ->, ultra thick] (a1.0) -- node[below=2.5mm]{\Huge 1} (1.8, 9.3);
                \path[draw, dashed, -, ultra thick] (a2.270) -- node [left=2.5mm, near end]{\Huge 2} (5.5, 4.5);
                \path[draw, dashed, ->, ultra thick] (5.5,4.5) -- (6, 4.5);
                \path[draw, dashed, -, ultra thick] (6.5, 4) -- (6.5, 3.7);
                \path[draw, dashed,->, ultra thick] (6.5, 3.7) -- node[above=2.5mm]{\Huge 3} (8.8, 3.7);
                \path[draw, dashed, -, ultra thick] (9, 9.5) -- (9.3, 9.5);
                \path[draw, dashed, -> , ultra thick] (9.3, 9.5) -- node[left=2.5mm, near end]{\Huge 3} (9.3, 4.2);
                \path[draw, dashed, -, ultra thick] (2.3, 8.8) -- node[left=2.5mm]{\Huge 4} (2.3, 1.0);
                \path[draw, dashed, ->, ultra thick] (2.3, 1.0) -- (8.3, 1.0);
                
                % \draw[draw=grey] (0, 0) grid (10, 10); 
            \end{tikzpicture}
            }
        \caption{}
        \label{fig:three_agent_buttons}
    \end{subfigure}
    \begin{subfigure}{0.65\linewidth}
        \resizebox{0.9\textwidth}{!}{
            \begin{tikzpicture}
                \node[state, initial above] (ui) {$\rminitstate$};
                \path (ui.0)+(\hDist, 0.0cm) node (u1) [state] {$\rmstate_1$};
                \path (u1.0)+(\hDist, 0.0cm) node (u2) [state] {$\rmstate_2$};
                \path (u2.90)+(\hDist, \vDist) node (u3) [state] {$\rmstate_3$};
                \path (u2.270)+(\hDist, -\vDist) node (u4) [state] {$\rmstate_4$};
                \path (u3.270)+(\hDist, -\vDist) node (u5) [state] {$\rmstate_5$};
                \path (u5.0)+(\hDist, 0.0cm) node (u6) [state] {$\rmstate_6$};
                \path (u6.270)+(0.0cm, -\vDist) node (u7) [state, accepting] {$\rmstatefinal$};
                
                \draw (ui) -> node[above, yshift=0.1cm] {$\yellowButton$} (u1);
                \draw (u1) edge node[above, yshift=0.1cm] {$\greenButton$} (u2);
                \draw (u2) edge node[below, xshift=0.2cm] {$\agentTwoRedButton$} (u3);
                \draw (u2) edge node[above, xshift=0.2cm] {$\agentThreeRedButton$} (u4);
                \draw (u3) edge node[below, xshift=-0.2cm] {$\agentThreeRedButton$} (u5);
                \draw (u4) edge node[above, xshift=-0.2cm] {$\agentTwoRedButton$} (u5);
                \draw (u3) edge[bend right] node[left, xshift=-0.1cm] {$\agentTwoNotRedButton$} (u2);
                \draw (u4) edge[bend left] node[left, xshift=-0.1cm] {$\agentThreeNotRedButton$} (u2);
                \draw (u5) edge node[above, yshift=0.1cm] {$\redButton$} (u6);
                \draw (u5) edge[bend right] node[right, xshift=0.15cm] {$\agentThreeNotRedButton$} (u3);
                \draw (u5) edge[bend left] node[right, xshift=0.1cm] {$\agentTwoNotRedButton$} (u4);
                \draw (u6) edge node[left, xshift=-0.1cm] {$\goal$} (u7);
            \end{tikzpicture}}
        \caption{}
        \label{fig:three_buttons_rm}
    \end{subfigure}
    \caption{Illustration of the \threebuttons grid (a) and a reward machine modeling the task's structure (b) \cite{Neary_Xu_Wu_Topcu_2021}.}
\end{figure*}

At time $t$, the agent keeps a trace $\ltrace_t\in(2^\propositions)^+$, and observes state $s_t\in\mdpstates$.
%label $\proplabel=\mdplfunc(s_t)$.
%and $\mdpterm(\ltrace_t, s_t)$ (i.e., whether history $(\ltrace_t, s_t)$ is terminal). If the history is non-terminal, 
The agent chooses the action to execute with its policy $a = \policy(\ltrace_t, s_t)\in\mdpactions$, and the environment transitions to state $s_{t+1}\sim\mdptransition(\cdot\mid s_t, a_t)$. As a result, the agent observes a new state $s_{t+1}$ and a new label $\proplabel_{t+1} = l(s_t, s_{t+1})$, and gets the reward $r_{t+1}=\mdpreward(\ltrace_{t+1},s_{t+1})$. The trace $\ltrace_{t+1}=\ltrace_t \oplus \proplabel_{t+1}$ is updated with the new label.

In this work, we focus on the \emph{goal-conditioned} RL problem \citep{kaelbling1993learning} where the agent's task is to reach a goal state as rapidly as possible.
We thus distinguish between two types of traces: \emph{goal} traces and \emph{incomplete} traces. A trace $\ltrace_t$ is said to be a \emph{goal} trace if the task's goal has been achieved before $t$ otherwise we say that the trace is \emph{incomplete}. 
Formally, the MDP includes a \emph{termination function} $\mdpterm:(2^{\propositions})^+\times\mdpstates\to \{\bot, \top\}$ indicating that the trace $\ltrace_t$ at time $t$ is an \emph{incomplete} trace if $\mdpterm(\ltrace_t, s_t) = \bot$ or a \emph{goal} trace if $\mdpterm(\ltrace_t, s_t) = \top$. 
We assume the reward is 1 for goal traces and 0 otherwise.
For ease of notation, we will refer to a goal trace as $\lgtrace_t$ and an incomplete trace as $\litrace_t$. 
Note that we assume the agent-environment interaction stops when the task's goal is achieved. This is without loss of generality as it is equivalent to having a final absorbing state with a null reward associated and no label returned by the labeling function $\mdplfunc$. 
%
%In the $T$-episodic case, the maximum length of a trace is therefore $T$.

The single-agent framework above can be extended to the collaborative \emph{multi-agent} setting as a Markov game. A cooperative \emph{Markov game} of $\numagents$ agents is a tuple $\markovgame=\langle N, \mgamestate, \mgameinitstate,\mgameactions,\mdptransition,\mdpreward,\mdpterm, T,\mdpdisc,\propositions,\mdplfunc\rangle$, where $\mgamestate=\mdpstates_1\times \cdots \times\mdpstates_\numagents$ is the set of joint states,  $\mgameinitstate\in\mgamestate$ is a joint initial state, $\mgameactions=\mdpactions_1\times \cdots \times \mdpactions_\numagents$ is the set of joint actions, $\mdptransition:\mgamestate \times \mgameactions \to \Delta(\mgamestate)$ is a joint transition function, $\mdpreward:(\mgamestate\times\mgameactions)^+\times\mgamestate\to\mathbb{R}$ is the collective reward function, $\mdpterm:(2^\propositions)^+ \times \mgamestate \to \{\bot, \top\}$ is the collective termination function, and $\mdplfunc:\mgamestate\times\mgamestate\to 2^\propositions$ is an event labeling function. $T$, $\mdpdisc$, and $\propositions$ are defined as for MDPs. 
Like \citet{Neary_Xu_Wu_Topcu_2021}, we assume each agent $A_i$'s dynamics are independently governed by local transition functions $\mdptransition_i:\mdpstates_i\times\mdpactions_i\to\Delta(\mdpstates_i)$, hence the joint transition function is $\mdptransition(\bm{s'}\mid\bm{s},\bm{a})=\Pi_{i=1}^\numagents\mdptransition(s'_i\mid s_i, a_i)$ for all $\bm{s},\bm{s'}\in\mgamestate$ and $\bm{a}\in\mgameactions$. The objective is to find a team policy $\policy:(2^\propositions)^+\times\mgamestate\to\mgameactions$ mapping pairs of traces and joint states to joint actions that maximizes the expected cumulative collective reward.

\begin{example}
    We use the \threebuttons task \cite{Neary_Xu_Wu_Topcu_2021}, illustrated in Figure~\ref{fig:three_agent_buttons}, as a running example. The task consists of three agents ($A_1$, $A_2$, and $A_3$) that must cooperate for agent $A_1$ to get to the $\goal$ position. To achieve this, agents must open doors that prevent other agents from progressing by pushing buttons $\yellowButton$, $\greenButton$, and $\redButton$. The button of a given color opens the door of the same color and, two agents are required to push $\redButton$ at once hence involving synchronization. Once a door has been opened, it remains open until the end of the episode. The order in which high-level actions must be performed is shown in the figure: (1)~$A_1$ pushes $\yellowButton$, (2)~$A_2$ pushes $\greenButton$, (3)~$A_2$ and $A_3$ push $\redButton$, and (4)~$A_1$ reaches $\goal$. 
    We can formulate this problem using the RL framework by defining a shared reward function returning $1$ when the $Goal$ position is reached. States solely consist of the position of the agents in the environment. The \emph{proposition set} in this task is $\propositions=\{\redButton,\allowbreak\yellowButton,\allowbreak\greenButton,\allowbreak\agentTwoNotRedButton,\allowbreak\agentTwoRedButton,\allowbreak\agentThreeNotRedButton,\allowbreak\agentThreeRedButton,\allowbreak\goal\}$, where (i)~$\redButton$, $\yellowButton$ and $\greenButton$ indicate that the red, yellow and green buttons have been pushed respectively, (ii)~$A_i^{R_B}$ (resp.~$A_i^{\neg R_B}$), where $i\in\{2,3\}$, indicates that agent $A_i$ is (resp.~has stopped) pushing the red button $\redButton$, and (iii)~$\goal$ indicates that the goal position has been reached. 
    A minimal \emph{goal trace} is $\langle\{\},\allowbreak \{\yellowButton\},\allowbreak \{\greenButton\},\allowbreak\{\agentTwoRedButton\},\allowbreak\{\agentThreeRedButton\},\allowbreak\{\redButton\},\allowbreak\{\goal\}\rangle$.
    \label{example:three_buttons_intro}
\end{example}

\subsection{Reward Machines}
We here introduce reward machines, the formalism we use to express decomposable (multi-agent) tasks. 
%
%First, we provide the basic definitions. Second, we describe the RL algorithm used to exploit the structure of reward machines. Third, we explain how reward machines have been used in the multi-agent case in the past.

\subsubsection{Definitions}
\hfill

\noindent
A \emph{reward machine} (RM) \cite{Icarte_Klassen_Valenzano_McIlraith_2018,IcarteKVM22} is a finite-state machine that represents the reward function of an RL task. 
Formally, an RM is a tuple $\rmname =\langle\rmstates,\propositions, \rminitstate, \rmstatefinal, \rmdeltau, \rmdeltar\rangle$ where $\rmstates$ is a set of states; $\propositions$ is a set of propositions; $\rminitstate\in\rmstates$ is the initial state; $\rmstatefinal\in\rmstates$ is the final state; $\rmdeltau:\rmstates \times 2^\propositions \to \rmstates$ is a state-transition function such that $\rmdeltau(u,\proplabel)$ is the state that results from observing label $\proplabel\in 2^\propositions$ in state $u\in\rmstates$; and $\rmdeltar:\rmstates \times \rmstates \to \mathbb{R}$ is a reward-transition function such that $\rmdeltar(u,u')$ is the reward obtained for transitioning from state $u\in \rmstates$ to $u'\in\rmstates$. We assume that (i)~there are no outgoing transitions from $\rmstatefinal$, and (ii)~$\rmdeltar(\rmstate,\rmstate')=1$ if $\rmstate'=\rmstatefinal$ and 0 otherwise (following the reward assumption in Section~\ref{sec:back_rl}).
Given a trace $\ltrace=\langle\proplabel_0, \ldots, \proplabel_n \rangle$, a \emph{traversal} (or run) is a sequence of RM states $\langle v_0, \ldots, v_{n+1} \rangle$ such that $v_0=u_0$, and $v_{i+1}=\rmdeltau(v_i,\proplabel_i)$ for $i\in[1,n]$. 
%
%However, unlike other approaches \cite{Icarte_Klassen_Valenzano_McIlraith_2018,XuGAMNT020, Furelos_Blanco_Law_Jonsson_Broda_Russo_2021}, we adopt the definition by \citet{Neary_Xu_Wu_Topcu_2021}, which processes  each label's propositions sequentially in no particular order (i.e., the state transition function takes a single proposition at a time).

%{\color{red}\textbf{Daniel:} Apparently the multi-agent work decomposes labels into single events and passes them to the RM. That is, if I observed $\{a,b\}$ (i.e., $a$ and $b$ at once), then we would first apply a transition using $a$ then another using $b$. Therefore, the transition function does not work like mine or other RMs in the literature and takes a single proposition at a time. Question: how have you modeled the transition function in ILASP? Does it learn formulas, a single symbol, ...? How do you use the transition function in your work? We need to know these things to be accurate. We will need to rewrite the transition function and the traversal if what I've described doesn't match your work.}
%\leo{We assume that there is always one event at a time. Therefore the transition of the RM are associated with a single event} {\color{red}I need more details. While we assume there's only one event at a time, I'm not sure whether this is reflected in the ILASP learning task (as I say above). I used to learn formulas: are you still learning formulas}

Ideally, RMs are such that (i)~the cross product $\mdpstates \times \rmstates$ of their states with the MDP's make the reward and termination functions Markovian, and (ii)~traversals for goal traces end in the final state $\rmstatefinal$ and traversals for incomplete traces do not.

\begin{example}
    Figure~\ref{fig:three_buttons_rm} shows an RM for the \threebuttons task. For simplicity, edges are labeled using the single proposition that triggers the transitions instead of sets. Note that the goal trace introduced in Example~\ref{example:three_buttons_intro} ends in the final state $\rmstatefinal$.
\end{example}

\subsubsection{RL Algorithm}
\label{sec:back_rl_algo}
\hfill

\noindent
The Q-learning for RMs (QRM) algorithm~\cite{Icarte_Klassen_Valenzano_McIlraith_2018,IcarteKVM22} exploits the task structure modeled through RMs. QRM learns a Q-function $\qfunc_\rmstate$ for each state $\rmstate\in\rmstates$ in the RM. Given an experience tuple $\langle s,a,s'\rangle$, a Q-function $\qfunc_\rmstate$ is updated as follows:
\begin{align*}
    \qfunc_\rmstate(s,a) = \qfunc_\rmstate(s,a)+\alpha\left(\rmdeltar(\rmstate,\rmstate') + \mdpdisc\max_{a'}\qfunc_{\rmstate'}(s',a') - \qfunc_\rmstate(s,a) \right),
\end{align*}
where $\rmstate'=\rmdeltau(\rmstate,\mdplfunc(s, s'))$. All Q-functions (or a subset of them) are updated at each step using the same experience tuple in a counterfactual manner.
%
%hence, independently of the function being currently used by the agent, any other function will benefit from what the agent is experiencing. 
In the tabular case, QRM is guaranteed to converge to an optimal policy.

%\begin{itemize}
    %\item Memory
    %\item Task structure and decomposition
    %\item Some non-Markovian reward
    %\item interpretability
    %\item trace types
%\end{itemize}

%{\color{red}
%Some things to clarify
%\begin{itemize}
%    \item Do we want to use the reward transition function?
%    \item Assume 0/1 reward function
%\end{itemize}
%}

%Single agent RM learning: \cite{Furelos_Blanco_Law_Jonsson_Broda_Russo_2021}

%Algo?

\subsubsection{Multi-Agent Decomposition}
\label{sec:task_decomposition}
\hfill

\noindent
\citet{Neary_Xu_Wu_Topcu_2021} recently proposed to decompose the RM for a multi-agent task into several RMs (one per agent) executed in parallel. The task decomposition into individual and independently learnable sub-tasks addresses the problem of non-stationarity inherent in the multi-agent setting. The use of RMs gives the high-level structure of the task each agent must solve. In this setting, each agent $A_i$ has its own RM $\rmname_i$, local state space $\mdpstates_i$ (e.g., it solely observes its position in the grid), and propositions $\propositions_i$ such that $\propositions=\bigcup_i\propositions_i$. Besides, instead of having a single opaque labeling function, each agent $A_i$ employs its own labeling function $\mdplfunc_i: \mdpstates_i \times \mdpstates_i \to 2^{\propositions_i}$. Each labeling function is assumed to return at most one proposition per agent per timestep, and they should together output the same label as the global labeling function $\mdplfunc$.

Given a global RM $\rmname$ modeling the structure of the task at hand (e.g.,~that in Figure~\ref{fig:three_buttons_rm}) and each agent's proposition set $\propositions_i$, \citet{Neary_Xu_Wu_Topcu_2021} propose a method for deriving each agent's RM $\rmname_i$ by projecting $\rmname$ onto $\propositions_i$. We refer the reader to \citet{Neary_Xu_Wu_Topcu_2021} for a description of the projection mechanism and its guarantees since we focus on learning these individual RMs instead (see Section~\ref{sec:learning_rms}).

\begin{figure}
    \begin{subfigure}{\columnwidth}
        \centering
        \begin{tikzpicture}
            % Horizontal format with colorful symbols
            \node[state, initial] (ui) {$\rminitstate^1$};
            \path (ui.0)+(\hDist, 0.0cm) node (u1) [state] {$\rmstate_1^1$};
            \path (u1.0)+(\hDist, 0.0cm) node (u2) [state] {$\rmstate_2^1$};
            \path (u2.0)+(\hDist, 0.0cm) node (u3) [state, accepting] {$\rmstatefinal^1$};
            
            \draw (ui) -> node[below, yshift=-0.1cm] {$\yellowButton$} (u1);
            \draw (u1) edge node[below, yshift=-0.1cm] {$\redButton$} (u2);
            \draw (u2) edge node[below, yshift=-0.1cm] {$\goal$} (u3);
        \end{tikzpicture}
        \caption{RM for $A_1$.}
        \label{fig:rm_a1}
    \end{subfigure}
    
    \begin{subfigure}{\columnwidth}
        \centering
        \begin{tikzpicture}
            \node[state, initial] (ui) {$\rminitstate^2$};
            \path (ui.0)+(\hDist, 0.0cm) node (u1) [state] {$\rmstate_1^2$};
            \path (u1.0)+(\hDist, 0.0cm) node (u2) [state] {$\rmstate_2^2$};
            \path (u2.0)+(\hDist, 0.0cm) node (u3) [state] {$\rmstate_3^2$};
            \path (u3.90)+(0.0cm, \vDist) node (u4) [state, accepting] {$\rmstatefinal^2$};
            
            \draw (ui) -> node[below, yshift=-0.1cm] {$\yellowButton$} (u1);
            \draw (u1) edge node[below, yshift=-0.1cm] {$\greenButton$} (u2);
            \draw (u3) edge node[right, xshift=0.1cm] {$\redButton$} (u4);
            
            \draw (u2) edge node[below, yshift=-0.1cm] {$\agentTwoRedButton$} (u3);
            \draw (u3) edge[bend right] node[above, yshift=0.1cm] {$\agentTwoNotRedButton$} (u2);
        \end{tikzpicture}
        \caption{RM for $A_2$.}
        \label{fig:rm_a2}
    \end{subfigure}
    
    \begin{subfigure}{\columnwidth}
        \centering
        \begin{tikzpicture}
            \node[state, initial left] (ui) {$\rminitstate^3$};
            \path (ui.0)+(\hDist, 0.0cm) node (u1) [state] {$\rmstate_1^3$};
            \path (u1.0)+(\hDist, 0.0cm) node (u2) [state] {$\rmstate_2^3$};
            \path (u2.0)+(\hDist, 0.0cm) node (u3) [state, accepting] {$\rmstatefinal^3$};
            
            \draw (ui) -> node[below, yshift=-0.1cm] {$\greenButton$} (u1);
            \draw (u2) edge node[below, yshift=-0.1cm] {$\redButton$} (u3);
            
            \draw (u1) edge node[below, yshift=-0.1cm] {$\agentThreeRedButton$} (u2);
            
            \draw (u2) edge[bend right] node[above, yshift=0.1cm] {$\agentThreeNotRedButton$} (u1);
        \end{tikzpicture}
        \caption{RM for $A_3$.}
         \label{fig:rm_a3}
    \end{subfigure}
    \caption{RMs for each of the agents in \threebuttons \cite{Neary_Xu_Wu_Topcu_2021}.}
    \label{fig:decomposed_rms}
\end{figure}
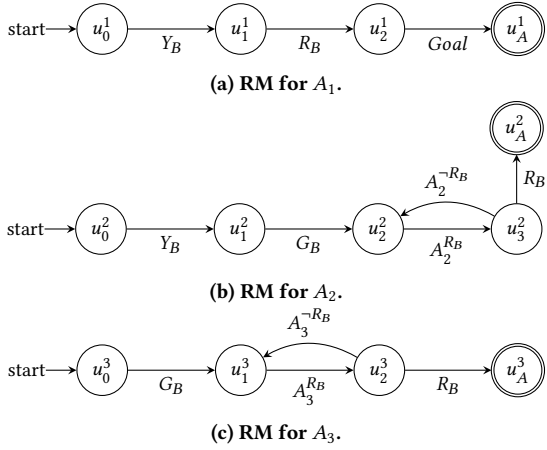

\begin{example}
    Given the local proposition sets $\propositions_1=\{\yellowButton,\redButton,\allowbreak\goal\}$, $\propositions_2=\{\yellowButton,\greenButton,\agentTwoRedButton,\agentTwoNotRedButton,\redButton\}$ and $\propositions_3=\{\greenButton,\agentThreeRedButton,\agentThreeNotRedButton,\redButton\}$, Figure~\ref{fig:decomposed_rms} shows the RMs that result from applying \citet{Neary_Xu_Wu_Topcu_2021} projection algorithm.
\end{example}

% To learn a decentralised policy, 
%
\citet{Neary_Xu_Wu_Topcu_2021} extend QRM and propose a decentralised training approach to train each agent in isolation (i.e., in the absence of their teammates) exploiting their respective RMs; crucially, the learnt policies should work when all agents interact simultaneously with the world.
In both the individual and team settings, agents must synchronize whenever a shared proposition is observed (i.e., a proposition in the proposition set of two or more agents). 
Specifically, an agent should check with all its teammates whether their labeling functions also returned the same proposition. In the individual setting, given that the rest of the team is not actually acting, synchronization is simulated with a fixed probability of occurrence.

%- use a specific MDP for each agent task? I guess this is more for learning the RM.

\section{Learning Reward Machines in Cooperative Multi-Agent Tasks}
\label{sec:learning_rms}
Decomposing a task using RMs is a promising approach to solving collaborative problems where coordination and synchronization among agents are required, as described in Section~\ref{sec:task_decomposition}. 
In well-understood problems, such as the \threebuttons task, one can manually engineer the RM encoding the sequence of sub-tasks needed to achieve the goal. 
However, this becomes a challenge for more complex problems, where the structure of the task is unknown.
In fact, the human-provided RM introduces a strong bias in the learning mechanism of the agents and can become adversarial if the human intuition about the structure of the task is incorrect.
To alleviate this issue, we argue for learning each of the RMs automatically from traces collected via exploration instead of handcrafting them.
In the following paragraphs, we describe two approaches to accomplish this objective.

\subsection{Learn a Global Reward Machine}
A naive approach to learning each agent's RM consists of two steps. First, a global RM $\rmname$ is learnt from traces where each label is the union of each agent's label. Different approaches \citep{Furelos_Blanco_Law_Jonsson_Broda_Russo_2021,IcarteWKVCM19,XuGAMNT020} could be applied in this situation.
The learnt RM is then decomposed into one per agent by projecting it onto the local proposition set $\propositions_i$ of each agent $A_i$ \cite{Neary_Xu_Wu_Topcu_2021}.

Despite its simplicity, this method is prone to scale poorly. It has been observed that learning a minimal RM (i.e., an RM with the fewest states) from traces becomes significantly more difficult as the number of constituent states grows \cite{Furelos_Blanco_Law_Jonsson_Broda_Russo_2021}. Indeed, the problem of learning a minimal finite-state machine from a set of examples is NP-complete \citep{Gold78}. Intuitively, the more agents interacting with the world, the larger the global RM will become, hence impacting performance and making the learning task more difficult. Besides, having multiple agents potentially increases the number of observable propositions as well as the length of the traces, hence increasing the complexity of the problem further.

\subsection{Learn Individual Reward Machines}
We propose to decompose the global task into sub-tasks, each of which can be independently solved by one of the agents. Note that even though an agent is assigned a sub-task, it does not have any information about its structure nor how to solve it. This is in fact what the proposed approach intends to tackle: learning the RM encoding the structure of the sub-task. This should be simpler than learning a global RM since the number of constituent states becomes smaller.

Given a set of $\numagents$ agents, we assume that the global task can be decomposed into $\numagents$ sub-task MDPs. Similarly to \citet{Neary_Xu_Wu_Topcu_2021}, we assume each agent's MDP has its own state space $\mdpstates_i$, action space $\mdpactions_i$, transition function $\mdptransition_i$, and termination function $\mdpterm_i$. In addition, each agent has its own labeling function $\mdplfunc_i$; hence, the agent may only observe a subset $\propositions_i \subseteq \propositions$.  Note that the termination function is particular to each agent; for instance, agent $A_2$ in \threebuttons will complete its sub-task by pressing the green button, then the red button until it observes the signal $\redButton$ indicating that $A_3$ is also pressing it. Ideally, the parallel composition of the learned RMs captures the collective goal.

We propose an algorithm that \emph{interleaves} the induction of the RMs from a collection of label traces and the learning of the associated Q-functions, akin to that by \citet{Furelos_Blanco_Law_Jonsson_Broda_Russo_2021} in the single agent setting. The decentralised learning of the Q-functions is performed using the method by \citet{Neary_Xu_Wu_Topcu_2021}, briefly outlined in Section~\ref{sec:task_decomposition}. The induction of the RMs is done using a state-of-the-art inductive logic programming system called ILASP~\cite{ILASP_system}, which learns the transition function of an RM as a set of logic rules from example traces observed by the agent.

\begin{example}
    The leftmost edge in Figure~\ref{fig:rm_a1} corresponds to the rule $\delta(u^1_0, u^1_1, \mathtt{T}) \codeif \mathtt{prop}(Y_B, \mathtt{T}).$ The $\mathtt{\delta(X,Y,T)}$ predicate expresses that the transition from \texttt{X} to \texttt{Y} is satisfied at time \texttt{T}, while the $\mathtt{prop(P, T)}$ predicate indicates that proposition \texttt{P} is observed at time \texttt{T}. The rule as a whole expresses that the transition from $u^1_0$ to $u^1_1$ is satisfied at time \texttt{T} if $Y_B$ is observed at that time. The traces provided to RM learner are expressed as sets of \texttt{prop} facts; for instance, the trace $\langle\{Y_B\}, \{R_B\}, \{Goal\}\rangle$ is mapped into $\{\mathtt{prop}(Y_B, 0).~\allowbreak\mathtt{prop}(R_B, 1).~\allowbreak\mathtt{prop}(Goal, 2).\}$.
\end{example}

\RestyleAlgo{ruled}
\begin{algorithm}
\caption{Multi-Agent QRM with RM Learning}\label{alg:Algo}
\SetKwFunction{InitializeQ}{InitializeQ}
\SetKwFunction{InitializeRM}{InitializeRM}
\SetKwFunction{GetAction}{GetAction}
\SetKwFunction{EnvStep}{EnvStep}
\SetKwFunction{GetNextRMState}{GetNextRMState}
\SetKwFunction{UpdateQ}{UpdateQ}
\SetKwFunction{LearnRM}{LearnRM}
\SetKwFunction{TrainAgent}{TrainAgent}
\SetKwFunction{GenerateIncompleteTraces}{GenerateIncompleteTraces}
\SetKwFunction{RMLearner}{RMLearner}
\SetKwProg{Procedure}{Procedure}{}

\For{$i=1$ \KwTo $\numagents$}{ \label{alg_line:init_start}
    $\rmstates^i \leftarrow \{ \rminitstate^i, \rmstatefinal^i \}$\\
    $\rmname^{i} \leftarrow$ \InitializeRM{$\rmstates^i$}\\
    $q^{i} \leftarrow$ \InitializeQ{$\rmname^i$} \\
    $\traceset^i_{\top} \leftarrow \emptyset$, 
    $\traceset^i_{\bot} \leftarrow \emptyset$ \label{alg_line:init_end}
}

\For{$n=1$ \KwTo $NumEpisodes$}{
    $t \leftarrow 0$ \\
    \For{$i=1$ \KwTo $\numagents$}{ \label{alg_line:episode_init_start}
        $u^{i}_{t} \leftarrow \rminitstate^i$, $s^{i}_t \leftarrow s^{i}_I$, $\ltrace^i_t \leftarrow \emptyset$, $done^i \leftarrow \bot$ \label{alg_line:episode_init_end}
    }
    \While{$\exists j \in N \textnormal{~such that~} done^j = \bot$}{\label{alg_line:learning_start}
        \For{$i=1$ \KwTo $\numagents$}{
            \TrainAgent{$done^{i}, s^{i}_{t}, u^{i}_t, q^{i}, \ltrace^{i}_t, \traceset^{i}_G, \traceset^{i}_I, M^{i}, \rmstates^{i}$}
        }
        $t \leftarrow t+1$
    }\label{alg_line:learning_end}
}
\hfill \\
\Procedure{\TrainAgent{$done, s_t, \rmstate_t, q, \ltrace_t, \gtraceset, \itraceset, \rmname, \rmstates$}}{
    \If{$done = \bot$}{
        $a \leftarrow$ \GetAction{$q_{u_{t}}, s_t$} \label{alg_line:rl_step_start} \\
        $s_{t+1}, isGoalAchieved \leftarrow$ \EnvStep{$s_t, a$} \label{alg_line:perform_step} \\
        $\proplabel_{t+1} \leftarrow \mdplfunc(s_{t}, s_{t+1})$ \label{alg_line:obtain_label} \\
        $\ltrace_{t+1} \leftarrow \ltrace_{t} \oplus \proplabel_{t+1}$ \label{alg_line:trace_extension} \\
        $r, u_{t+1} \leftarrow$ \GetNextRMState{$\rmname, u_{t}, \proplabel_{t+1}$} \label{alg_line:obtain_next_rm_state} \\
        $q_{u_{t}} \leftarrow$ \UpdateQ{$q_{u_{t}}, s_t, a, r, s_{t+1}, q_{u_{t+1}}$} \label{alg_line:q_update} \\
        \For{$u \in \rmstates \setminus \{ u_{t} \}$}{
            $r, u' \leftarrow$ \GetNextRMState{$\rmname, u, \proplabel_{t+1}$} \\
            $q_{u} \leftarrow$ \UpdateQ{$q_{u}, s_t, a, r, s_{t+1}, q_{u'}$} \label{alg_line:counterfactual_q_update} \\
        }\label{alg_line:rl_step_end}
        \hfill \\ \label{alg_line:rm_learning_start}
        \If{$isGoalAchieved$}{ \label{alg_line:goal_trace}
            $\gtraceset \leftarrow \gtraceset \cup \{\ltrace_{t+1}\}$\\
            $\itraceset \leftarrow \itraceset \cup \GenerateIncompleteTraces(\ltrace_{t+1})$\\
            $\rmname_{new} \leftarrow$ \LearnRM{$\rmstates, \gtraceset, \itraceset$} \\
            $done \leftarrow \top$
        }
        \ElseIf{$u_{t+1} = \rmstatefinal$}{ \label{alg_line:rm_done}
            $\itraceset \leftarrow \itraceset \cup \{\ltrace_{t+1}\}$\\
            $\rmname_{new} \leftarrow$ \LearnRM{$\rmstates,\gtraceset, \itraceset$} \\
            $done \leftarrow \top$
        }
        \ElseIf{$t+1 = T$}{ \label{alg_line:max_steps}
            $\itraceset \leftarrow \itraceset \cup \{\ltrace_{t+1}\}$ \\
            $done \leftarrow \top$
        }
        \hfill \\
        \If{$\rmname \neq \rmname_{new}$}{
            $\rmname \leftarrow \rmname_{new}$ \\
            $q \leftarrow$ \InitializeQ{$\rmname$} \label{alg_line:rm_reset_q}
        }\label{alg_line:rm_learning_end}
    }
}
\hfill \\
\Procedure{\LearnRM{$\rmstates, \gtraceset, \itraceset$}}{ \label{alg_line:learn_rm_start}
    $M \leftarrow \RMLearner(\rmstates, \gtraceset, \itraceset)$\label{alg_line:ilasp_first_call}\\
    \While{$M = \bot$}{
        $\rmstates \leftarrow \rmstates \cup \{\rmstate_{|\rmstates|-1}\}$\\
        $M \leftarrow \RMLearner(\rmstates, \gtraceset, \itraceset)$
    }
    \Return $M$ \label{alg_line:learn_rm_end}
}
\end{algorithm}

Algorithm~\ref{alg:Algo} shows the pseudocode describing how the reinforcement and RM learning processes are interleaved. The algorithm starts initializing for each agent $A_i$: the set of states $\rmstates^i$ of the RM, the RM $\rmname^i$ itself, the associated Q-functions $\qfunc^i$, and the sets of example goal ($\traceset^i_{\top}$) and incomplete ($\traceset^i_{\bot}$) traces (\texttt{l.\ref{alg_line:init_start}-\ref{alg_line:init_end}}). Note that, initially, each agent's RM solely consists of two unconnected states: the initial and final states (i.e., the agent loops in the initial state forever). Next, $NumEpisodes$ are run for each agent. Before running any steps, each environment is reset to the initial state $s^i_I$, each RM $\rmname^i$ is reset to its initial state $\rminitstate^i$, each agent's trace $\ltrace^i$ is empty, and we indicate that no agents have yet completed their episodes (\texttt{l.\ref{alg_line:episode_init_start}-\ref{alg_line:episode_init_end}}). Then, while there are agents that have not yet completed their episodes (\texttt{l.\ref{alg_line:learning_start}}), a training step is performed for each of the agents, which we describe in the following paragraphs. Note that 
%(i)~an external counter $t$ for the number of steps is kept (as we will see later, this is leveraged inside each agent's training step), and (ii)~
we show a sequential implementation of the approach but it could be performed in parallel to improve performance.

% The algorithm considers that each sub-task is trained in isolation in a separate instance of the environment where only one agent operates. 

The $\texttt{TrainAgent}$ routine shows how the reinforcement learning and RM learning processes are interleaved for a given agent $A_i$ operating in its own environment. From \texttt{l.\ref{alg_line:rl_step_start}} to \texttt{l.\ref{alg_line:rl_step_end}}, the steps for QRM (i.e., the RL steps) are performed. The agent first selects the next action $a$ to execute in
its current state $s_t$ given the Q-function $\qfunc_{\rmstate_t}$ associated with the current RM state $\rmstate_t$. The action is then applied, and the agent observes the next state $s_{t+1}$ and whether it has achieved its goal (\texttt{l.\ref{alg_line:perform_step}}). The next label $\proplabel_{t+1}$ is then determined (\texttt{l.\ref{alg_line:obtain_label}}) and used to (i)~extend the episode trace $\ltrace_t$ (\texttt{l.\ref{alg_line:trace_extension}}), and (ii)~obtain reward $\mdpreward$ and the next RM state $\rmstate_{t+1}$ (\texttt{l.\ref{alg_line:obtain_next_rm_state}}). The Q-function $\qfunc$ is updated for both the RM state $\rmstate_t$ the agent is in at time $t$ (\texttt{l.\ref{alg_line:q_update}}) and in a counterfactual manner for all the other states of the RM (\texttt{l.\ref{alg_line:counterfactual_q_update}}).

The learning of the RMs occurs from \texttt{l.\ref{alg_line:goal_trace}} to \texttt{l.\ref{alg_line:rm_learning_end}}. Remember that there are two sets of example traces for each agent: one for goal traces $\gtraceset$ and one for incomplete traces $\itraceset$. Crucially, the learnt RMs must be such that (i)~traversals for goal traces end in the final state, and (ii)~traversals for incomplete traces do not end in the final state. Given the trace $\ltrace_{t+1}$ and the RM state $\rmstate_{t+1}$ at time $t+1$, the example sets are updated in three cases:
\begin{enumerate}
    \item If $\ltrace_{t+1}$ is a goal trace (\texttt{l.\ref{alg_line:goal_trace}}), $\ltrace_{t+1}$ is added to $\gtraceset$. We consider as incomplete all sub-traces of $\ltrace_{\top}$: $\{\langle\proplabel_0, \ldots, \proplabel_{k} \rangle; \forall k \in [0, \ldots, |\ltrace_{\top}|-1] \}$ (see Example~\ref{ex:incomplete_examples_generation}). Note that if a sub-trace was a goal trace, the episode would have ended before. This optimization enables capturing incomplete traces faster; that is, it prevents waiting for them to appear as counterexamples (see next case).
    
    % We motivate the latter operation in Section~\ref{sec:optimizations}. 
    
    %By already considering all incomplete subtraces at this point, we avoid delaying their observation until later in the form of counterexamples (see next case).
    \item If $\ltrace_{t+1}$ is an incomplete trace and $\rmstate_{t+1}$ is the final state of the RM (\texttt{l.\ref{alg_line:rm_done}}), then $\ltrace_{t+1}$ is a counterexample since reaching the final state of the RM should be associated with completing the task. In this case, we add $\ltrace_{t+1}$ to $\itraceset$.
    \item If $\ltrace_{t+1}$ is an incomplete trace and the maximum number of steps per episode $T$ has been reached (\texttt{l.\ref{alg_line:max_steps}}), we add $\ltrace_{t+1}$ to $\itraceset$.
\end{enumerate}
In the first two cases, a new RM is learnt once the example sets have been updated. The \texttt{LearnRM} routine (\texttt{l.\ref{alg_line:learn_rm_start}-\ref{alg_line:learn_rm_end}}) finds the RM with the fewest states that covers the example traces. The RM learner (here ILASP) is initially called using the current set of RM states $\rmstates$ (\texttt{l.\ref{alg_line:ilasp_first_call}}); then, if the RM learner finds no solution covering the provided set of examples, the number of states is increased by 1. This approach guarantees that the RMs learnt at the end of the process are \emph{minimal} (i.e., consist of the fewest possible states)~\cite{Furelos_Blanco_Law_Jonsson_Broda_Russo_2021}. Finally, like \citet{Furelos_Blanco_Law_Jonsson_Broda_Russo_2021}, we reset the Q-functions associated with an RM when it is relearned; importantly, the Q-functions depend on the RM structure and, hence, they are not easy to keep when the RM changes (\texttt{l.\ref{alg_line:rm_reset_q}}). In all three enumerated cases, the episode is interrupted.

%Then, at some point we should mention what guarantees stated in that work are not being granted here (I guess, e.g. all benefits that come from the projection).

%Mention assumptions on the labeling functions of each agent (make sure that any needed relationship with previous work is mentioned in the corresponding background section).

%\subsubsection{Optimizations}
%\label{sec:optimizations}
%\hfill

%\noindent
%To improve the learning efficiency we apply the subsequent optimizations to the set of examples used by the RM learner:
%\begin{itemize}
    % \item Like \citet{Furelos_Blanco_Law_Jonsson_Broda_Russo_2021}, we start learning the RM only after having observed a goal trace.
    %\item For every goal trace $\ltrace_{\top}$, we consider as incomplete all the sub-traces of $\ltrace_{\top}$: $\{\langle\proplabel_0, \ldots, \proplabel_{k} \rangle; \forall k \in [0, \ldots, |\ltrace_{\top}|-1] \}$ (see Example~\ref{ex:incomplete_examples_generation}). Note that if a sub-trace was a goal trace, the episode would have ended before. This optimization enables capturing incomplete traces faster and prevents waiting for them to appear as counterexamples.
    %\item We make sure to keep distinct traces in our set of examples of goal and incomplete traces. 
    %
    %It helps avoid a growing set of examples not bringing new information to the RM learner.
%\end{itemize}

\begin{example}
    Given the sub-task assigned to the agent $A_3$ in the \threebuttons environment described in Figure~\ref{fig:rm_a3}, we consider a valid goal trace $\ltrace = \langle\{\},\allowbreak \{\greenButton\},\allowbreak \{\agentThreeRedButton\},\allowbreak\{\agentThreeNotRedButton\},\allowbreak \{\agentThreeRedButton\},\allowbreak\{\redButton\}\rangle$. The set of generated incomplete traces is:
    \begin{itemize}
        \item[] $\langle\{\}\rangle$,
        \item[] $\langle\{\},\allowbreak \{\greenButton\}\rangle$,
        \item[] $\langle\{\},\allowbreak \{\greenButton\},\allowbreak \{\agentThreeRedButton\}\rangle$,
        \item[] $\langle\{\},\allowbreak \{\greenButton\},\allowbreak \{\agentThreeRedButton\},\allowbreak\{\agentThreeNotRedButton\}\rangle$,
        \item[] $\langle\{\},\allowbreak \{\greenButton\},\allowbreak \{\agentThreeRedButton\},\allowbreak\{\agentThreeNotRedButton\},\allowbreak \{\agentThreeRedButton\}\rangle$
    \end{itemize}
    \label{ex:incomplete_examples_generation}
\end{example}

We have described in this section a method to learn the RM associated with each sub-task. In the next section, we evaluate how this approach performs on two collaborative tasks.

\section{Experiments}

We evaluate our approach by looking at two metrics as the training progresses: (1)~the collective reward obtained by the team of agents and (2)~the number of steps required for the team to achieve the goal. The first metric indicates whether the team learns how to solve the task by getting a reward of $1$ upon completion. The second one estimates the quality of the solution, i.e. the lower the number of steps the more efficient the agents. Although the training is performed in isolation for each agent, the results presented here were evaluated with all the agents acting together in a shared environment. Additionally, we inspect the RM learnt by each agent and perform a qualitative assessment of their quality.
The results presented show the mean value and the $95\%$-confidence interval of the metrics evaluated over $5$ different random seeds in two grid-based environments of size $7$x$7$.
We ran the experiments on an \textsc{EC2 c5.2xlarge} instance on \textit{AWS} using \textit{Python} $3.7.0$. We use the state-of-the-art inductive logic programming system \textit{ILASP} v$4.2.0$ \cite{ILASP_system} to learn the RM from the set of traces with a timeout set to $1h$.

\subsection{\threebuttons Task}

\begin{figure}
\centering
\captionsetup{justification=centering}
\captionsetup[sub]{font=small}
\hspace{\fill}
\begin{subfigure}[t]{0.5\textwidth}\centering
    \includegraphics[width=0.9\textwidth]{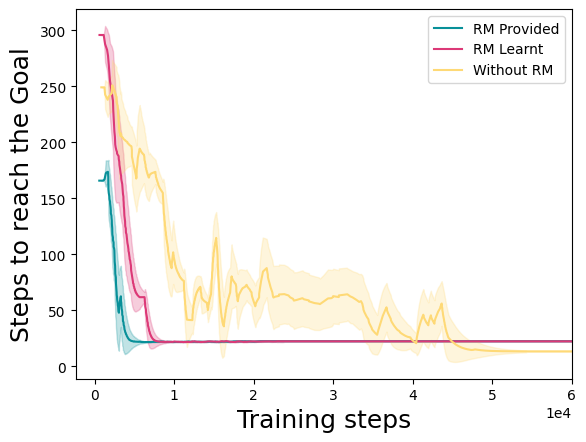}
    \caption{Number of Steps to reach the Goal}
    \label{fig:threebuttons_steps}
\end{subfigure}
\hfill
\begin{subfigure}[t]{0.5\textwidth}\centering
    \includegraphics[width=0.9\textwidth]{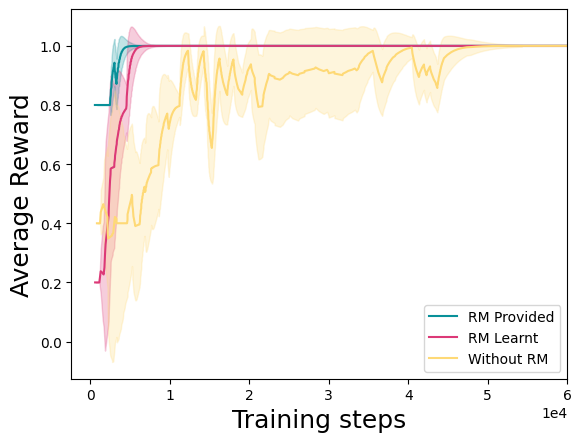}
    \caption{Average Reward}
    \label{fig:threebuttons_reward}
\end{subfigure}
\hspace{\fill}
\caption{Comparison between handcrafted RMs (RM Provided) and our approach learning the RMs from traces (RM Learnt) in the \threebuttons environment.}
\label{fig:three_buttons_performances}
\end{figure}

We evaluate our approach in the \threebuttons task described in Example~\ref{example:three_buttons_intro}. The number of steps required to complete the task and the collective reward obtained by the team of agents throughout training are shown in Figure~\ref{fig:three_buttons_performances}. The \emph{green} curve shows the performance of the team of agents when the local RMs are manually engineered and provided. In \emph{pink}, we show the performance of the team when the RMs are learnt. Finally in \emph{yellow}, we show the performance of the team when each agent learn independently to perform their task without using the RM framework\footnote{This is equivalent of using a 2-State RM, reaching the final state only when the task is completed.}. We observe that regardless of whether the local RMs are handcrafted or learned, the agents learn policies that, when combined, lead to the completion of the global task. Indeed, the collective reward obtained converges to the maximum reward indicating that all the agents have learnt to execute their sub-tasks and coordinate in order to achieve the common goal. Learning the RMs also helps speed up the learning for each of the agent, significantly reducing the number of training iterations required to learn to achieve the global task. We note however that the policy learnt without the RM framework requires less steps than when the RMs are used.

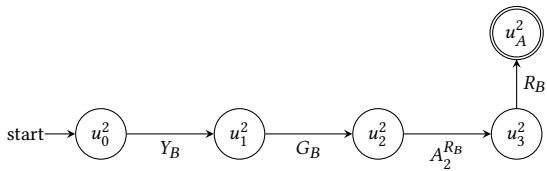
\begin{figure}
    \centering
    \begin{tikzpicture}
        \node[state, initial] (ui) {$\rminitstate^2$};
        \path (ui.0)+(\hDist, 0.0cm) node (u1) [state] {$\rmstate_1^2$};
        \path (u1.0)+(\hDist, 0.0cm) node (u2) [state] {$\rmstate_2^2$};
        \path (u2.0)+(\hDist, 0.0cm) node (u3) [state] {$\rmstate_3^2$};
        \path (u3.90)+(0.0cm, \vDist) node (u4) [state, accepting] {$\rmstatefinal^2$};
        
        \draw (ui) -> node[below, yshift=-0.1cm] {$\yellowButton$} (u1);
        \draw (u1) edge node[below, yshift=-0.1cm] {$\greenButton$} (u2);
        \draw (u3) edge node[right, xshift=0.1cm] {$\redButton$} (u4);
        
        \draw (u2) edge node[below, yshift=-0.1cm] {$\agentTwoRedButton$} (u3);
    \end{tikzpicture}
    \caption{Learnt RM for $A_2$.}
    \label{fig:learnt_rm_a2}
\end{figure}

We present in Figure~\ref{fig:learnt_rm_a2} the RM learnt by $A_2$ using our interleaved learning approach. We compare it to the ``true'' RM shown in Figure~\ref{fig:rm_a2}. The back transition from $\rmstate^2_3$ to $\rmstate^2_2$ associated with the label $\agentTwoNotRedButton$ is missing from the learnt RM. This can be explained by the fact that this transition does not contribute towards the goal. When deriving the minimal RM from the set of labels, this one in particular is not deemed important and is thus ignored by the RM learner.

Learning the global RM was also attempted in this environment. Unfortunately, learning the minimal RM for the size of this problem is hard and the RM learner timed out after the $1h$ limit.

\subsection{\rendezvous Task}

\begin{figure*}
    \captionsetup{justification=centering}
    \centering
    \begin{subfigure}{0.325\textwidth}
        \centering
        \resizebox{0.8\columnwidth}{!}{
            \begin{tikzpicture}[scale=1.0]
                \filldraw[fill=black!0!white] (0,0) rectangle (10,10);
            
                \filldraw[fill=green, opacity=0.6, draw=black] (5, 5) circle (0.75cm);
                
                \node at (5, 5) (rdv) {\Huge $RDV$};
                
                \node[anchor=center] at (1.0, 9.0) (g) {\Huge $\goal_1$};
                \node[anchor=center] at (9.0, 1.0) (g) {\Huge $\goal_2$};
                \node[anchor=center] at (2.0, 1.0) (a1) {\Huge $A_1$};
                \node[anchor=center] at (9.0, 9.0) (a2) {\Huge $A_2$};
                
                \draw[draw=black, very thick] (0,0) rectangle (10, 10);
                
                \path[draw, dashed, -, ultra thick] (2.0, 1.5) -- node[left=2.5mm]{\Huge 1} (2.0, 4.75);
                \path[draw, dashed, ->, ultra thick] (2.0, 4.75) -- (4.0, 4.75);
                \path[draw, dashed, -, ultra thick] (4.0, 5.25) -- node[above=2.5mm]{\Huge 2} (1.0, 5.25);
                \path[draw, dashed, ->, ultra thick] (1.0, 5.25) -- (1.0, 8.5);
    
                \path[draw, dashed, -, ultra thick] (9.0, 8.5) -- node[right=2.5mm]{\Huge 1} (9.0, 5.25);
                \path[draw, dashed, ->, ultra thick] (9.0, 5.25) -- (6.0, 5.25);
                \path[draw, dashed, -, ultra thick] (6.0, 4.75) -- node[below=2.5mm]{\Huge 2} (9.0, 4.75);
                \path[draw, dashed, ->, ultra thick] (9.0, 4.75) -- (9.0, 1.5);
                
                % \draw[draw=grey] (0, 0) grid (10, 10); 
            \end{tikzpicture}
        }
        \caption{}
        \label{fig:rendezvous_task_example}
    \end{subfigure}
    \begin{subfigure}{0.65\linewidth}
        \centering
        \resizebox{0.9\textwidth}{!}{
            \begin{tikzpicture}
                \node[state, initial above] (ui) {$\rminitstate$};
                \path (ui.270)+(\hDist, \vDist) node (u1) [state] {$\rmstate_1$};
                \path (ui.90)+(\hDist, -\vDist) node (u2) [state] {$\rmstate_2$};
                \path (u1.90)+(\hDist, -\vDist) node (u3) [state] {$\rmstate_3$};
                \path (u3.0)+(\hDist, 0.0cm) node (u4) [state] {$\rmstate_4$};
                \path (u4.270)+(\hDist, \vDist) node (u6) [state] {$\rmstate_6$};
                \path (u4.90)+(\hDist, -\vDist) node (u5) [state] {$\rmstate_5$};
                \path (u6.90)+(\hDist, -\vDist) node (u7) [state, accepting] {$\rmstatefinal$};
                
                \draw (ui) -> node[above, yshift=0.1cm, xshift=0.1cm] {$R_1$} (u2);
                \draw (u2) edge[bend left] node[left, yshift=-0.1cm] {$\neg R_1$} (ui);
                \draw (ui) -> node[below, yshift=-0.1cm, xshift=0.1cm] {$R_2$} (u1);
                \draw (u1) edge[bend right] node[left, yshift=0.2cm, xshift=0.2cm] {$\neg R_2$} (ui);
                \draw (u1) -> node[below, yshift=-0.1cm, xshift=-0.1cm] {$R_1$} (u3);
                \draw (u2) -> node[above, yshift=0.1cm, xshift=-0.1cm] {$R_2$} (u3);
                \draw (u3) edge[bend left] node[left, yshift=-0.2cm, xshift=0.4cm] {$\neg R_2$} (u2);
                \draw (u3) edge[bend right] node[left, yshift=0.1cm, xshift=0.4cm] {$\neg R_1$} (u1);
                \draw (u3) -> node[above, yshift=0.1cm] {$R$} (u4);
                \draw (u4) -> node[above, yshift=0.1cm, xshift=-0.1cm] {$G_1$} (u6);
                \draw (u4) -> node[below, yshift=-0.1cm, xshift=-0.1cm] {$G_2$} (u5);
                \draw (u5) -> node[below, yshift=-0.1cm, xshift=0.1cm] {$G_1$} (u7);
                \draw (u6) -> node[above, yshift=0.1cm, xshift=0.1cm] {$G_2$} (u7);
            \end{tikzpicture}}
        \caption{}
        \label{fig:rendezvous_rm}
    \end{subfigure}
    \caption{Example of the \rendezvous task where $2$ agents must meet on the RDV point (green) before reaching their goal state $G1$ and $G2$ for agents $A1$ and $A2$ respectively.}
\end{figure*}
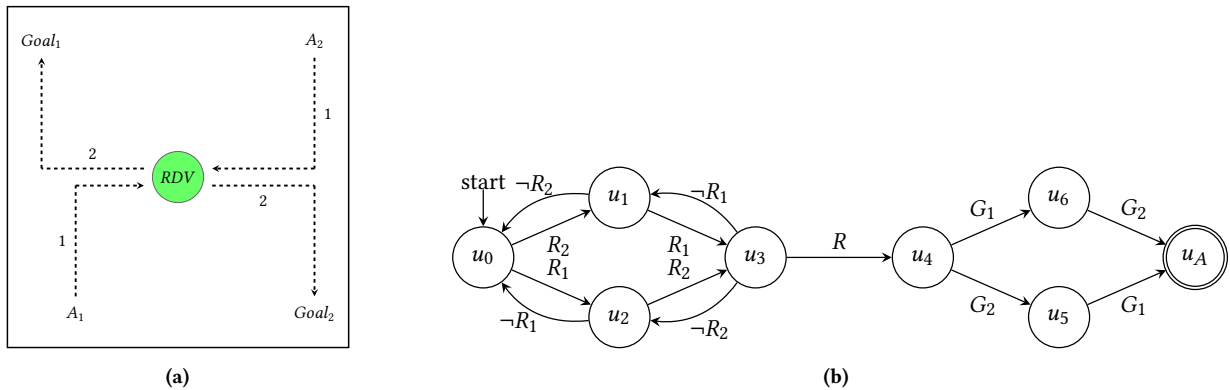

The second environment in which we evaluate our approach is the $2$-agent \rendezvous task \cite{Neary_Xu_Wu_Topcu_2021} presented in Figure~\ref{fig:rendezvous_task_example}. This task is composed of $2$ agents acting in a grid with the ability to go \textsc{up}, \textsc{down}, \textsc{left}, \textsc{right} or \textsc{do nothing}. The task requires the agents to move in the environment to first meet in an $RDV$ location, i.e, all agents need to simultaneously be in that location for at least one timestep. Then each of the agents must reach their individual goal location for the global task to be completed.

\begin{figure}
\centering
\captionsetup{justification=centering}
\captionsetup[sub]{font=small}
\hspace{\fill}
\begin{subfigure}[t]{0.5\textwidth}\centering
    \includegraphics[width=0.9\textwidth]{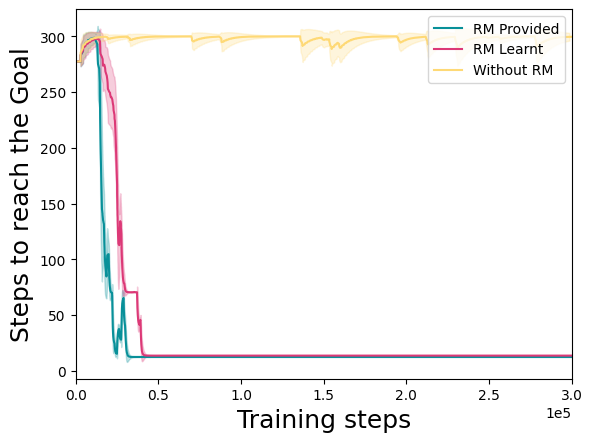}
    \caption{Number of Steps to reach the Goal}
    \label{fig:rendezvous_steps}
\end{subfigure}
\hfill
\begin{subfigure}[t]{0.5\textwidth}\centering
    \includegraphics[width=0.9\textwidth]{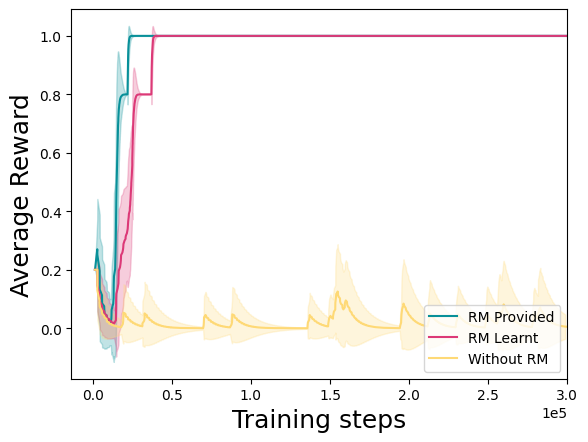}
    \caption{Average Reward}
    \label{fig:rendezvous_reward}
\end{subfigure}
\hspace{\fill}
\caption{Comparison between handcrafted RMs (RM Provided) and our approach learning the RMs from traces (RM Learnt) in the \rendezvous environment.}
\label{fig:rendezvous_performances}
\end{figure}

We present in Figure~\ref{fig:rendezvous_performances} the number of steps (\ref{fig:rendezvous_steps}) to complete the task and the collective reward received by the team of agents (\ref{fig:rendezvous_reward}) throughout training. In \emph{green} we show the performance when the RMs are known a priori and provided to the agents. This scenario, where we have perfect knowledge about the structure of each sub-tasks, is used as an upper bound for the results of our approach. In \emph{pink}, we show the performances of our approach where the local RMs are learnt. The \emph{yellow} curve shows the performance of the team when each agent learns to perform their individual task without the RM construct.

The collective reward obtained while interleaving the learning of the policies and the learning of the local RMs converges to the maximum reward after a few iterations. In this task, the RM framework is shown to be crucial to help the team solve the task in a timely manner. Indeed, without the RMs (either known a priori or learnt), the team of agents do not succeed at solving the task.

We also attempted to learn the global RM directly but the task could not be solved (i.e., a collective goal trace was not observed). The collective reward remained null throughout the training that we manually stopped after $1e6$ timesteps.

\section{Related Work}
Since the recent introduction of reward machines as a way to model non-Markovian rewards \cite{Icarte_Klassen_Valenzano_McIlraith_2018,IcarteKVM22}, several lines of research have based their work on this concept (or similar finite-state machines).
Most of the work has focused on how to derive them from logic specifications~\cite{CamachoIKVM19} or demonstrations~\cite{CamachoVZJIK21}, as well as learning them using different methods~\cite{IcarteWKVCM19,Furelos_Blanco_Law_Jonsson_Broda_Russo_2021,Gaon_Brafman_2020,XuGAMNT020,HasanbeigJAMK21,FurelosBlancoLJBR22,ChristoffersenLTM20}.

In this work, based on the multi-agent with RMs work by \citet{Neary_Xu_Wu_Topcu_2021}, the RMs of the different agents are executed in parallel. Recent works have considered other ways of composing RMs, such as merging the state and reward transition functions \cite{DeGiacomo20} or arranging them hierarchically by enabling RMs to call each other \cite{FurelosBlancoLJBR22}. The latter is a natural way of extending this work since agents could share subroutines in the form of callable RMs.

The policy learning algorithm for exploiting RMs we employ is an extension of the QRM algorithm (see Sections~\ref{sec:back_rl_algo}-\ref{sec:task_decomposition}). However, algorithms based on making decisions at two hierarchical levels~\cite{Icarte_Klassen_Valenzano_McIlraith_2018,Furelos_Blanco_Law_Jonsson_Broda_Russo_2021} or more \cite{FurelosBlancoLJBR22} have been proposed. In the simplest case (i.e., two levels), the agent first decides which transition to satisfy from a given RM state, and then decides which low-level actions to take to satisfy that transition. For example, if agent $A_2$ gets to state $u^2_3$ in Figure~\ref{fig:rm_a2}, the decisions are: (1)~whether to satisfy the transition labeled $\agentTwoNotRedButton$ or the one labeled $\redButton$, and (2)~given the chosen transition, act towards satisfying it. Unlike QRM, this method is not guaranteed to learn optimal policies; however, it enables lower-level sub-tasks to be reused while QRM does not. Note that the Q-functions learned by QRM depend on the structure of the whole RM and, hence, we learn them from scratch every time a new RM is induced. While other methods attempt to leverage the previously learned Q-functions \cite{XuGAMNT020}, the hierarchical approach outlined here does not need to relearn all functions each time a new RM is induced.

In the multi-agent community, the use of finite-state machines for task decomposition in collaborative settings has been studied in \citep{7040356, elsefy2020task} where a top-down approach is adopted to construct sub-machines from a global task.
Unlike our work, these methods focus on the decomposition of the task but not on how to execute it. The algorithm presented in this paper interleaves both the decomposition and the learning of the policies associated with each state of the RM.

The MARL community studying the collaborative setting has proposed different approaches to decompose the task among all the different agents. For example, QMIX \cite{rashid2020monotonic} tackles the credit assignment problem in MARL assuming that the team's Q-function can be factorized, and performs a value-iteration method to centrally train policies that can be executed in a decentralised fashion. While this approach has shown impressive empirical results, its interpretation is more difficult than understanding the structure of the task using RM.

The use of RMs in the MARL literature is starting to emerge. The work of \citet{Neary_Xu_Wu_Topcu_2021} was the first to propose the use of RMs for task decomposition among multiple agents. As already mentioned in this paper, this work assumes that the structure of the task is known `a priori', which is often untrue.
With a different purpose, \citet{Dann0A0T22} propose to use RMs to help an agent predict the next actions the other agents in the system will perform. Instead of modeling the other agents' plan (or program), the authors argue that RMs provide a more permissive way to accommodate for variations in the behaviour of the other agent by providing a higher-level mechanism to specify the structure of the task. In this work as well, the RMs are pre-defined and provided to the agent.

Finally, the work of \citet{DistSPECTRL} uses a finite-state machine called `task monitor', which is built from temporal logic specifications that encode the reward signal of the agents in the system. A task monitor is akin to an RM with some subtle differences like the use of registers for memory.

\section{Conclusion and Future Work}

Dividing a problem into smaller parts that are easier to solve is a natural approach performed instinctively by humans. It is also particularly suited for the multi-agent setting where all the available resources can be utilized to solve the problem or when a complementary set of skills is required.

In this work, we extend the approach by \citet{Neary_Xu_Wu_Topcu_2021}, who leverage RMs for task decomposition in a multi-agent setting. We propose a novel approach where the RMs are learnt instead of manually engineered. 
The method presented in this paper interleaves the learning of the RM from the traces collected by the agent and the learning of the set of policies used to achieve the task's goal. Learning the RMs associated with each sub-task not only helps deal with non-Markovian rewards but also provides a more interpretable way to understand the structure of the task.
We show experimentally in the \threebuttons and the \rendezvous environments that our approach converges to the maximum reward a team of agents can obtain by collaborating to achieve a goal. Finally, we validated that the learnt RM corresponds indeed to the RM needed to achieve each sub-task.

While our approach lifts the assumption about knowing the structure of the task `a priori', a set of challenges remain and could be the object of further work:
\begin{enumerate}
    \item A labeling function is assumed to exist for each of the agents and be known by them. In other words, each agent knows the labels relevant to its task and any labels used for synchronization with the other team members.
    \item If the labeling functions are noisy, agents may fail to synchronize and hence not be able to progress in their sub-tasks. In this scenario, the completion of the cooperative task is compromised. Defining or learning a communication protocol among the agents could be the object of future work.
    \item Our approach assumes that the global task has been automatically divided into several tasks, each to be completed by one of the agents. Dynamically creating these sub-tasks and assigning them to each of the agents is crucial to having more autonomous agents and would make an interesting topic for further research.
    %given in the form of a termination function; that is, the global task is not
    %\item To learn the local RM and the associated RL policies, the agent must be rewarded once it has accomplished its sub-task. However, the completion of the sub-task is often different from the completion of the global task (e.g., in the \threebuttons task). Each individual therefore requires their own termination function $\mdpterm_i$ to be known, implying that the high-level decomposition of the global task is also known.
    \item We leverage task decomposition to learn simpler RMs that when run in parallel represent a more complex global RM. Task decomposition can be driven further through task hierarchies, which could be represented through hierarchies of RMs~\cite{FurelosBlancoLJBR22}. Intuitively, each RM in the hierarchy should be simpler to learn and enable reusability and further parallelization among the different agents.
\end{enumerate}
%
%In the \threebuttons example, the labeling function of the agent $A_1$ only returns the sub-set of propositions $\propositions_1 = \{\redButton,\allowbreak\yellowButton,\allowbreak \goal\}$ instead of the full proposition set of the environment $\propositions=\{\redButton,\allowbreak\yellowButton,\allowbreak\greenButton,\allowbreak\agentTwoNotRedButton,\allowbreak\agentTwoRedButton,\allowbreak\agentThreeNotRedButton,\allowbreak\agentThreeRedButton,\allowbreak\goal\}$. This considerably reduces the length of the traces, hence making the RM easier to learn. 
%

The use of RMs in the multi-agent context is still in its infancy and there are several exciting avenues for future work. 
We have demonstrated in this paper that it is possible to learn them under certain assumptions that could be relaxed in future work.

\balance

\begin{acks}
Research was sponsored by the Army Research Laboratory and was accomplished under Cooperative Agreement Number W911NF-22-2-0243. The views and conclusions contained in this document are those of the authors and should not be interpreted as representing the official policies, either expressed or implied, of the Army Research Laboratory or the U.S. Government. The U.S. Government is authorised to reproduce and distribute reprints for Government purposes notwithstanding any copyright notation herein. 
\end{acks}

\bibliographystyle{ACM-Reference-Format}
\bibliography{references}

%%%%%%%%%%%%%%%%%%%%%%%%%%%%%%%%%%%%%%%%%%%%%%%%%%%%%%%

\appendix

\end{document}